%% file: mnlp2020.tex
\pdfoutput=1

\documentclass[11pt]{article}

\usepackage[]{acl}

\usepackage{times}
\usepackage{latexsym}
\usepackage{multirow}
\usepackage{amsmath}

\usepackage{graphicx}
\usepackage{makecell}
\usepackage[T1]{fontenc}

\usepackage[utf8]{inputenc}

\usepackage{microtype}
\newcommand{\comment}[1]{}
\usepackage{booktabs}       
\usepackage{url}
\usepackage{adjustbox}

\title{Multilingual Open QA on the MIA Shared Task}

\author{
  Navya Yarrabelly\thanks{\hspace{4pt}Everyone Contributed Equally} 
  \hspace{2em} Saloni Mittal$^*$ 
  \hspace{2em} Ketan Todi$^*$ 
  \hspace{2em} Kimihiro Hasegawa$^*$ \\
  Language Technologies Institute \\
  \texttt{\{nyarrabe, salonim, ktodi, kimihiro\}@andrew.cmu.edu}
  }

\begin{document}
\maketitle

\input{sections/1-introduction}
\input{sections/2-relatedwork}
\input{sections/2-baseline}
\input{sections/3-methods}
\input{sections/4-experiment}
\input{sections/5-result}
\input{sections/6-conclusion}

\bibliographystyle{acl_natbib}
\bibliography{mnlp2020}

\end{document}

%% file: sections/1-introduction.tex
\section{Introduction}
\subsection {Cross-lingual information retrieval}
Cross-lingual information retrieval (CLIR) ~\cite{shi2021cross, asai2021one, jiang2020cross} for example,
can find relevant text in any language
such as English(high resource) or Telugu (low resource) even when the query is posed in a different, possibly low-resource, language. In
this work, we aim to develop useful CLIR models
for this constrained, yet important, setting where we do not require any kind of additional supervision or labelled data for retrieval task and hence can work effectively for low-resource languages. 

\par
We propose a simple and effective re-ranking
method for improving passage retrieval in
open question answering. The re-ranker re-scores retrieved passages with a zero-shot
multilingual question generation model, which is a pre-trained language model, to compute the probability of the input question in the target language  conditioned on a retrieved passage, which can be possibly in a different language. We evaluate our method in a completely zero shot setting and doesn't require any training. Thus the main advantage of our method is that our approach can be used to re-rank results obtained by any sparse retrieval methods like BM-25. This eliminates the need for obtaining expensive labelled corpus required for the retrieval tasks and hence can be used for low resource languages. 

\comment{
on top of any retrieval method (e.g. neural or
keyword-based), does not require any domainor task-specific training (and therefore is expected to generalize better to data distribution
shifts), and provides rich cross-attention between query and passage (i.e. it must explain
every token in the question)

}
\subsection{Multilingual Question Answering}

Multilingual question answering involves getting the answer in the same language as the question.
Until recently most models focused on single languages \cite{DBLP:journals/corr/RajpurkarZLL16, DBLP:journals/corr/abs-1909-07005}.
But with the recent multilingual models there has been a lot of progress of Multilingual QA with a myriad of datasets and models.
In this work we aim to improve upon the \cite{asai2021one} model.
As it can be seen from the baseline results there is almost a linear correlation between the number of training examples vs the test accuracy.
Hence we believe that increasing the number of examples for low resourced languages should help improve the performance.

The MIA competition had to tracks: constrained and unconstrained setup.
The first allowed data augmentation without adding new annotated QA pairs.
Combined with the high resource requirements to train such models, we decided to move forward with the constrained setup.
Korean and Bengali have the worst performance and least number of examples across all languages.
Hence to verify our hypothesis we augment the korean and bengali data by translating sentences from English to the respective language.

%% file: sections/2-relatedwork.tex
\section{Related Work}

\subsection{Multilingual QA}

One of the major bottleneck in case of multilingual QA is the lack of high quality training data unlike monolingual QA models which draw inspiration from translating SQUAD to a particular language.
One of the core focuses in multilingual QA has been to create high quality data while reducing annotation requirements.
Hence most multilingual QA models focus on (a) data augmentation techniques using neural machine translation and alignment methods or manually translating small subsets of data (b) annotate multilingual data, and lastly (c) use different transfer learning and modelling techniques to improve QA performance.

MLQA \cite{https://doi.org/10.48550/arxiv.1910.07475} adopted a simple approach of identifying passages with same content across multiple languages and then crowdsourced English question answer pairs annotation using the English passages.
The questions were then translated to other languages (six other languages) and an off-shelf aligner was used to align the passages and find answers in other languages.
But the major drawback of this approach was the error propagation from aligner.

XQUAD \cite{artetxe-etal-2020-cross} performed zero shot multilingual QA. 
They using MLM objective to learn language specific token embeddings and then used a QA model trained on english dataset to perform inference on the target language by replacing English token embeddings with target language embeddings.
Along with this they also experimented with adapters and language specific position embeddings.
One major drawback was the use of separate models for different languages.

TyDi-QA \cite{clark2020tydi} was the first open domain multilingual QA dataset not involving any kind of translation.
The annotators were asked to write questions followed by a google search on the question to retrieve the relevant wikipedia article.
The second task for the annotators involved identifying which passage in the article can answer the question, and extract the answer from that particular passage.
Otherwise that question can be masked as non answerable.
One major drawback of this approach was the poor performance of search engines on low resource datasets.

\subsection{Cross Lingual Retrieval}


Although open QA has been a highly researched area in the last few years, the work has largely been limited to systems exclusively in English. There has been little prior work in building a unified Cross Lingual Retrieval system for many-to-many open QA. 
Some attempts for cross-lingual retrieval in the past rely on machine-translation and monolingual retrieval in a specific language. 
XOR QA \cite{Asai2021XORQC} use train an NMT model to translate all queries to English and then retrieve in a monolingual way. For retrieval they use a BM25 \cite{INR-019}-based retrieval followed by dense neural re-ranking using DPR \cite{Karpukhin2020DensePR}.

The CORA paper \cite{asai2021one} is the first to propose a single cross-lingual retriever that does not rely on language-specific retrieval or translations.The multilingual retrieval module (mDPR) can produce dense embeddings of multilingual questions and passages in the space space and therefore, enable cross lingual retrieval by performing an inner product between the question and passage encodings. The mDPR is trained iteratively by automatic cross-lingual training data mining through Wikipedia language links and self-training based on model predictions. 

In this paper, we build upon the output results of the mDPR by treating it as a weak filter and adding another re-ranking component to the pipeline that processes the output passages from mDPR. Our approach uses a pre-trained mulltingual language model in a zero-shot setting to re-score the passages. Although, fundamentally both mDPR and our approach are build upon the same large pre-trained models, our approach is more expressive than the dense retrieval method used in mDPR as it incorporates rich token-level cross-attention between question and passages.

%% file: sections/2-baseline.tex
\section{Baseline}
\label{sec:baseline}


XOR-TYDI QA \cite{asai2020xor} aimed to alleviate the issue of non answerable questions in TYDI-QA.
The non answerable questions were first human translated to English, and then the English passage was used for extracting the correct answer using an English passage.
The English answer was then human translated back to the language of the question.
The 2 main advantages of this was (a) increase in training data for languages other then English, and (b) cross lingual answering where the model needs to generate answer in different language from English using English passages as the information source.
This dataset consists of 7 diverse languages as shown in \autoref{table:dataset}.

CORA \cite{asai2021one} adopted a 2 stage retrieve-then generate approach rather then using an extractive based.
CORA trained a multilingual DPR (mDPR) \cite{Karpukhin2020DensePR} where given a question the mDPR model needs to extract relevant passages from the corpora irrespective of the language of the passage.
The mBERT model was used as the base model for finetuning on the mDPR dataset.

\begin{figure}
    \centering
    \includegraphics[scale=0.3]{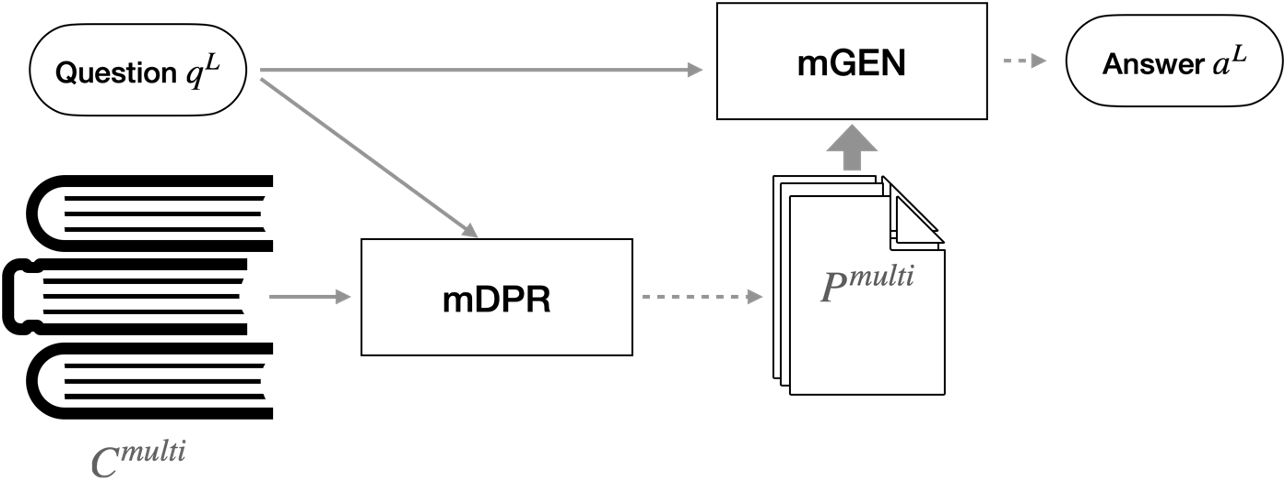}
    \caption{Block Diagram of our model architecture.}
    \label{fig:block_diagram}
\end{figure}

This is followed by an mT5 (multilingual T5) answer generation model which takes as input the question and a string consisting of the passages retrieved.
The mt5 model is then required to generate the answer in the same language as the question.
This is enforced by using a language token as BOS token for the model.  We show a basic block diagram of the architecture in \autoref{fig:block_diagram}.

\subsection{Multilingual Dense Passage Retriever (mDPR)}

Multilingual Dense Passage Retriever (mDPR) ~\cite{asai2021one} extends Dense Passage Retriever ~\cite{karpukhin2020dense} to a multilingual setting.   mDPR uses an iterative training approach to
fine-tune a pre-trained multilingual language model (e.g., mBERT) to encode
passages and questions separately. mDPR uses an iterative training approach to fine-tune a pre-trained multilingual language model (e.g., mBERT; Devlin et al., 2019) to encode
passages and questions separately.  A 768 dimensional passage representation is computed as follows.

\begin{equation}
    \mathbf{e}{p^{L}}=\mathrm{mBERT}{p}(p)
\end{equation}

where a passage p is a fixed-length sequence of tokens from multilingual
 mDPR model is trained for cross lingual retrieval task with the initial training data, which  is a combination of multilingual QA datasets: XORTYDI QA~\cite{asai2020xor} and TYDI QA~\cite{clark2020tydi}, and an English open QA dataset ~\cite{kwiatkowski2019natural}. They further use Wikipedia based data augmentation techniques to extend the passage set to other languages. 
 
\subsection{Multilingual Question Answering}

The top $k$ passages retrieved using mDPR $P^{multi}$ are concatenated together and passed as input to the mT5 model as shown in \autoref{eqn:answer_generation}.

\begin{equation}
    \label{eqn:answer_generation}
    a^{L} = mGEN(q^L, p^{multi})
\end{equation}

\begin{equation}
    \label{eqn:answer_generation_loss}
    P(a^{L}|q^{L}, P^{multi}) = \prod_{i}^{T} p(a^{L}_{i}|a^{L}_{<i},q^{L}, P^{multi})
\end{equation}

The answers are generated token-by-token by the mT5 model as shown in \autoref{eqn:answer_generation_loss}.
The major advantage of using a generation based approach is it helps the model to adapt to languages unseen during the training time.


%% file: sections/3-methods.tex
\section{Methods}

This section covers a detailed explanation of our two proposed approaches on improving retrieval and data augmentation for question answering part~\footnote{https://github.com/kimihiroh/11737}.

\subsection{Question-Generation based Re-ranking (QGPR)}

This approach is aimed at improving passage retrieval by using a simple re-ranking method on the top-k retrieved passages from the baseline mDPR model. The re-ranker is based on a zero-shot question generation model that uses a pre-trained multilingual language model (PMLM) to compute the likelihood of the given question conditioned on a retrieved passage from the mDPR. The question likelihood is used to re-score the passages before passing them to the answer generation model. This approach incorporates rich cross-attention between the query and passage tokens while forcing the model to explain every token in the question \cite{Sachan2022ImprovingPR}.

The re-ranking approach is based on zero-shot question generation model that uses a multilingual pre-trained model to score the probability of generating the question $q$ given the passage text $z$ i.e. $ p(q | z) $. More specifically, we estimate $ log p(q | z)$ by computing the average log likelihood of the question tokens conditioned on the passage:
\begin{equation}
    \log p\left(\boldsymbol{q} \mid \boldsymbol{z}\right)=\frac{1}{|\boldsymbol{q}|} \sum_{t} \log p\left(q_{t} \mid \boldsymbol{q}_{<t}, \boldsymbol{z} ; \Theta\right)    
\end{equation}

where $\Theta$ denotes the parameters of the PMLM and $|q|$ denotes the number of tokens in a question.

As part of this method, we performed four distinct experiments with two PMLMs namely, mT5~\cite{xue2020mt5} and mBART~\cite{liu2020multilingual}. The following details our 4 experiments.
\begin{enumerate}
    \item A zero-shot mT5 model used as a question-generation model. The absence of a language tag to control the target language in mT5 is a challenge for us as we are dealing with a cross-lingual dataset, where the language of the question is different from the passages. To alleviate this problem to some extent, we prompt the model to generate a question in a specific language x by appending the phrase, "Please generate a question in x for this passage", to the passage text. But a more challenging case for generative models is zero-shot cross-lingual span selection for question generation. Hence, we also tried using mBART, which is trained for machine translation tasks, for a slightly different objective where we tried to predict the question translation given a passage. 
    \item To further alleviate the cross-lingual generation problems for mT5, we translate the passage to the language of the question using an NMT model(mBART-50 ~\footnote{https://huggingface.co/facebook/mbart-large-50-many-to-many-mmt}) and repeat 1. The language of the question is known to us in the dataset.
    
    \item Similarly, we translate the question to the language of the passage using an NMT model and repeat 1. The language of the passage is detected using an off-the-shelf lang ID detection tool. We use langdetect\footnote{https://pypi.org/project/langdetect/} tool  for language detection. 
    
    \item A zero-shot mBART model is used as the question generation model. The main advantage of using mBART over mT5 is that the target language can be controlled by inputting a language tag to the model.
\end{enumerate}

\subsection{Machine-Translation-based Data Augmentation}
Aiming to improve the performance of the question answering part, or \textit{reader} part, we experiment an machine-translation-based data augmentation~\cite{lee-etal-2018-semi}. This is motivated by the fact that the low performance and the smaller number of Bengali and Korean examples. This approach consists of two steps: 1) We translate $(q^{en}, p^{en}, a^{en})$ into $(q^{L}, p^{L}, a^{L})$, where $q, p, a$ represents a question, a paragraph, and an answer, respectively. $en$ and $L$ means English and the target language $L$. 2) Then, among the translated examples, we only keep examples whose positive paragraphs $p^L_{pos}$ contain $a^L$ as is. This is to reduce noisy augmented data from a translation model.

%% file: sections/4-experiment.tex
\section{Experimental Details}
\input{tabels/dataset}
We evaluated  our approach on XOR-TYDI QA dataset~\cite{asai2020xor} which contains questions and related passages across 7 typologically diverse languages. Table~\ref{table:dataset} shows the statistics of the dataset.

\subsection{Question-Generation based Re-ranking}
We take the top-50 results obtained by mDPR ~\cite{asai2021one} model and re-rank these results with our approach. We do not use the mDPR retrieval score or rank of the passages in any manner while re-ranking the results. Thus, it can effectively be used for re-ranking the results obtained from any of the underlying retrieval models.

While generating the question in a target language l\textsubscript{a} given a passage in a language l\textsubscript{b}, the model has to deal with two complex tasks i.e 1. Generate for a new task. 2. Generate in a language which is different from that of the passage. 
\begin{itemize}

\item {\textbf{Question Generation with Cross Lingual Sources}}
We first used an mBART  ~\cite{liu2020multilingual} model to generate the question given the passage. We used a pretrained mbart-50 many-to-many checkpoint for generating the questions from a passage. Since mBART is trained with a denoising objective for translation tasks, we believe that it can generate  more accurate question translations for a given passage. Though we tried in a zero shot setting, we believe that a few shot method using contrastive loss can further boost the results. 
\item \textbf{Question Generation  with Monolingual Sources}
As explained in section 4, we translated passages to the same language as question and also translated questions to the same language as passage and used mT5 for zero shot question generation task with the translated data.

\end{itemize}
 
\par

\subsection{Machine-Translation-based Data Augmentation}
As the source of English tuples, ($q^{en}$, $p^{en}$, $a^{en}$), we use the retrieval results of mDPR, shared by the workshop organizer~\footnote{\url{https://github.com/mia-workshop/MIA-Shared-Task-2022}}. From the file, we take English examples. Then, considering the computational cost, the actual input sequence length to \textit{reader} model, and the number of examples of middle-resource languages, we translate three paragraphs for both positive and negative paragraphs for 5k examples. As the translation model, we use MBart~\cite{liu2020multilingual}~\footnote{\url{https://huggingface.co/docs/transformers/model_doc/mbart}}. As the \textit{reader} model, we use mT5-small~\cite{xue2020mt5}~\footnote{\url{https://huggingface.co/docs/transformers/model_doc/mt5}}. Due to the computational resource restriction, we limit the maximum input length as 600, instead of the default 1000 that the baseline result was calculated. We use one RTX 8000 with 48 GB for this experiment.

\begin{table*}[ht]
\centering
\begin{adjustbox}{width=\linewidth,center}
\begin{tabular}{l cc cc cc cc}
\toprule
& \multicolumn{2}{c}{ko} & \multicolumn{2}{c}{ja} & \multicolumn{2}{c}{fi} & \multicolumn{2}{c}{bn} \\
& P@5 & P@15 & P@5 & P@15 & P@5 & P@15 & P@5 & P@15 \\
\midrule
MDPR(baseline) & 0.298 & 0.646 & 0.07 & 0.126 & \textbf{0.622} & \textbf{1.135} & \textbf{0.518} & \textbf{0.971} \\
QGPR(m-BART) & \textbf{0.355} & \textbf{0.697} & \textbf{0.085} & \textbf{0.127} & 0.562 & 1.10 & 0.308 & 0.751 \\
QGPR(m-T5) & 0.218 & 0.615 & 0.037 & 0.097 & 0.334 & 0.809 & 0.338 & 0.87 \\
QGPR $L((tr(q_{l_a}) \rightarrow q_{l_b})|p_{l_b})$ & 0.1966 & 0.533 & 0.034 & 0.094 & 0.289 & 0.787 & 0.306 & 0.75 \\ 
QGPR $L(q_{l_a}|(tr(p_{l_b}) \rightarrow p_{l_a}))$ & 0.302 & 0.70 & 0.065 & 0.126 & 0.52 & 0.992 & 0.43 & 0.956 \\
\bottomrule
\end{tabular}
\end{adjustbox}
\caption{The following table shows the evaluation results for the Cross-Lingual Retrieval task. P@k refers to Positives@k for k=5,15.}
\label{tab:qgpr_result_precision}
\end{table*}

\begin{table*}[ht]
\centering
\begin{adjustbox}{width=\linewidth,center}
\begin{tabular}{l cc cc cc cc}
\toprule
& \multicolumn{2}{c}{ko} & \multicolumn{2}{c}{ja} & \multicolumn{2}{c}{fi} & \multicolumn{2}{c}{bn} \\
& R@5 & R@15 & R@5 & R@15 & R@5 & R@15 & R@5 & R@15 \\
\midrule
MDPR(baseline) & 12.1 & 21.4 & 5.5 & 9.1 & \textbf{29.0} & \textbf{41.3} & \textbf{18.8} & \textbf{30.6} \\
QGPR(m-BART) & \textbf{16.0} & \textbf{25.6} & \textbf{5.6} & 9.7 & 22.2 & 37.3 & 12.2 & 24.2 \\
QGPR(m-T5) & 8.02 & 20.8 & 2.28 & 6.64 & 13.87 & 29.04 & 10.77 & 24.7 \\
QGPR $L((\text{tr}(q_{l_a}) \rightarrow q_{l_b})|p_{l_b})$ & 6.92 & 17.15 & 2.18 & 6.3 & 10.0 & 24.53 & 7.7 & 19.2 \\ 
QGPR $L(q_{l_a}|(\text{tr}(p_{l_b}) \rightarrow p_{l_a}))$ & 12.0 & 22.8 & 4.57 & \textbf{10.67} & 24.54 & 38.6 & 13.43 & 28.0 \\
Gain & \textbf{+3.9} & \textbf{+4.2} & \textbf{+0.1} & \textbf{+0.6} & \textbf{-4.46} & \textbf{-4} & \textbf{-5.3} & \textbf{-1.4} \\
\bottomrule
\end{tabular}
\end{adjustbox}
\caption{The following table shows the evaluation results for the Cross-Lingual Retrieval task. R@K refers to Recall@K for k=5,15.}
\label{tab:qgpr_result_recall}
\end{table*}

%% file: tabels/dataset.tex
\begin{table}[ht]
\centering
\begin{adjustbox}{width=0.7\linewidth,center}
\begin{tabular}{l cc }
    \toprule
    Language & Training & Development \\
    \midrule
    English (en) & 91876 & --- \\
    Arabic (ar) & 15828 & 1387 \\
    Finnish (fi) & 7680 & 974\\
    Russian (ru) & 7349 & 1018 \\
    Japanese (ja) & 5527 & 693\\
    Telugu (te) & 5451 & 564 \\
    Bengali (bn) & 2428 & 490 \\
    Korean (ko) & 1856 & 473 \\
    \bottomrule
\end{tabular}
\end{adjustbox}
\caption{The number of examples in our target dataset, XOR-TyDi-QA.}
\label{table:dataset}
\end{table}

%% file: sections/5-result.tex
\section{Results and Analysis}

\subsection{Question-Generation based Re-ranking}
 We evaluated the efficacy of our re-ranking approach on different languages and compared them with the MDPR as the baseline. Tables~\ref{tab:qgpr_result_precision} and~\ref{tab:qgpr_result_recall} shows the results for our  Question Generation based Re-ranking experiments. \\
 \textbf{Metrics:} 
 We primarily used  the standard ranking evaluation metrics, which is Positives@K and Recall@K. Positives@K  is the number of positive passages appearing in the top K results and Recall@K is the  percentage of total positives appearing in the top-k results. For our experiments, we use the ground truth positives in the top-50 results obtained by mDPR as our total positives set. 
 \par
 Row 1 in  tables~\ref{tab:qgpr_result_precision} and~\ref{tab:qgpr_result_recall} shows the results of the baseline and rows 2-5 includes the results for our approaches and. Row 4 corresponds to QGPR with questions translated to the passage language and row 5 corresponds to QGPR with passages translated to the question language.  All our QGPR based  approaches are  completely unsupervised and do not require any additional training.

Our QGPR based approach using a zero-shot mBART model gave an performance improvement of 4\% in Recall@5 and Recall@15 over the baseline for Korean language, while giving a slight improvement of about 0.5\% for Japaneese language. However, we observed that it led to a performance degradation of about 4\% in Recall@15 for Finnish language and 1.4\% for Bengali language. 
While the mDPR model has seen the training data for the languages Korean, Japanese, Finnish and Bengali, our model still outperformed for high resource languages like Korean and Japanese. We hypothesize that the deteriorated performance on Bengali and Finnish is observed as they are low resource languages. As a result, mBART's generation quality is much worse when compared to Korean and Japanese.

The QGPR experiments based on mT5 performs poorly when compared to the baseline. This behavior was expected as there is no language tag provision to control the target language of the mT5 model. In its absence, the model is more likely to generate the question in the language of the input passage it is conditioned on. 

Precisely for this language mismatch problem between the question and passage, we carried out two other experiments where questions are translated in the language of the passage at the time of question generation and vice-e-versa. The success of this approach relies heavily on good quality translations obtained from the NMT model. In Table \ref{tab:qgpr_result_precision} and~\ref{tab:qgpr_result_recall} we can see that none of these approaches improve the baseline results. However, translating passages into question langauge (row 5 in the table) performs consistently better than the reverse direction for all languages in the Recall@k metrics. We hypothesize that even a small token-level error in translation of question can be highly detrimental to the model's retrieval performance. Since the question length is significantly smaller than the multi-sentence passages and contains important keywords in that short span, whereas the long passages give more room to accommodate translation errors.

\subsubsection {Ranking Shift Analysis - Cross Lingual vs Same Language}
To further analyse the ranking shift for passages in same language vs passages that are cross lingual, we computed Mean Reciprocal Rank (MRR) of the relevant passages and averaged over all the questions. A higher MRR means that the positive passages are ranked in the top positions and hence higher is better.
\par
Table ~\ref{tab:qgpr_mrr} shows the MRR for the Same setting (monolingual case where we compute MRR for positive passages that are in the same language as the questions) and Cross setting (crosslingual case where we compute MRR for positive passages that are in a language different  than the question) for our methods and compared them with the mDPR baseline. We evaluated the MRR of our re-ranking approach, Question Generation with passages translated to the same language as question,  on different languages and compared them with the mDPR baseline. 
\par
Our re-ranking method has improved MRR for cross lingual setting. While in general both the baseline and our re-ranking approaches have degraded performance in  retrieving cross lingual passages compared to the passages which are in same language as the question.

\begin{table*}[ht]
\centering
\begin{adjustbox}{width=0.9\linewidth,center}
\begin{tabular}{l cc cc cc cc }
    \toprule
    & \multicolumn{2}{c}{ko} & \multicolumn{2}{c}{ja} & \multicolumn{2}{c}{fi} & \multicolumn{2}{c}{bn} \\
    \cmidrule(r){2-3} \cmidrule(r){4-5} \cmidrule(r){6-7} \cmidrule(r){8-9}
     & Same & Cross & Same & Cross & Same & Cross & Same & Cross \\
    \midrule
     mDPR(baseline)& 0.226 & 0.0006 & 0.054 & 0.00176 & \textbf{0.401} & 0.05 & \textbf{0.348} & 0.0148 \\
    QGPR with passage translation & \textbf{0.260} & \textbf{0.0022} & 0.052 & \textbf{0.007} & 0.297 & \textbf{0.214} & 0.25 & \textbf{0.1938} \\
    \bottomrule
\end{tabular}
\end{adjustbox}
\caption{The following table lists Mean Reciprocal Ranking(MRR) of true passages for the retrieval task. Column Same refers to the passages that are in the same language as the question. Column Cross means that the passage is in a different language than the question.}

\label{tab:qgpr_mrr}
\end{table*}



\subsection{Machine-Translation-based Data Augmentation}

\textbf{Metrics}: We report the F1 scores of the baseline approach as compared to ours. 
In this case F1 score considers the predicted and golden answer as a list of tokenized strings.
It then calculates the precision and recall between the predicted and the golden list and then finally calculates the F1 score.

\input{tabels/result_mt_aug}

Table~\ref{table:results_mt_aug} shows the results of our machine-translation-based data augmentation approach. This is the F1 score. From the results, we do not see any consistent tendency. Especially, we do not see any noticeable improvements on Bengali and Korean, of which we added data. We see more than one point difference on Arabic, Japanese, and Telugu, although we made no manipulation on those data. We assume this is due to the training hyper-parameters, e.g., the number of training epochs, the random seeds and so on. 

On further analysing we found that out of 40 examples for which the augmented model gave correct results 10 of those examples only had english number as answers.
And similarly for telgu 7/15 examples only had english number as answers.
This improvement might have been due to increase in cross lingual information sources, as we increased questions in ko and te where many context were still in English.
Hence we believe that to further identify areas of improvement we might need to segregate the dataset in terms of whether the answer is numerical or non numerical.

One of the major reason for this inconsistent results is the maximum sequence length mT5 can process, i.e. 600 in our case.
Since we restrict the input length to 600 tokens, which means that on an average only 3 passages can be passed as input to the reader model.
So even after doing the data augmentation the number of questions increased for each of the 2 languages.
But due to the restriction on the input sequence length the context ($P^{multi}$) passed as input for these 2 languages (ko and bn) only had 1 positive translated context, the rest of the context might still have been in English, thereby reducing the affect of translated dataset.
Also since the R@5 scores are pretty low for the retrieval model this eventually puts a restriction on the maximum performance of the mGEN model.


%% file: tabels/result_mt_aug.tex
\begin{table*}[h]
\centering
\begin{adjustbox}{width=0.7\linewidth,center}
\begin{tabular}{l ccccccc }
    \toprule
     & ar & fi & ru & ja & te & bn & ko \\
    \midrule
     Baseline & 40.07 & 32.45 & 32.49 & 34.00 & 34.00 & 20.67 & 20.08 \\
     Ours & 41.48 & 32.39 & 32.26 & 32.70 & 35.11 & 19.82 & 20.29 \\
    \bottomrule
\end{tabular}
\end{adjustbox}
\caption{Result of Machine-Translation-based Data Augmentation}
\label{table:results_mt_aug}
\end{table*}

%% file: sections/6-conclusion.tex
\section{Conclusion}

In this report we explored methods for improving cross-lingual retrieval and answer generation for a multilingual open QA task. We show the efficacy of our method by reporting positive improvements over the baseline for the retrieval task. On the retrieval side, we proposed an unsupervised method for re-ranking retrieved passages from the baseline mDPR model. This approach can be applied
on top of any retrieval method and does not require any dataset-specific training (and therefore is expected to generalize better to data distribution shifts). Our positive results also validate our initial hypothesis that this approach is more expressive than mDPR-based dense retrieval as it forces the model to explain every token in the question. Given reasonably high resource languages like Korean and Japanese for which zero-shot pre-trained multilingual models are expected to do better, our re-ranking approach outperforms the mDPR baseline.

There was only a marginal improvement in the QA performance as a result of data augmentation.
This data augmentation approach had shown improvements in fixed gold passage QA settings.
We believe that increasing the input context size should help to improve performance on augmented dataset.



\section{Contribution of Team Members}
\begin{enumerate}
    \item Question-Generation based Re-ranking (QGPR) - Navya and Saloni
    \item Machine-Translation-based Data Augmentation - Kimi and Ketan
\end{enumerate}